\documentclass[10pt,twocolumn,letterpaper]{article}

\usepackage[pagenumbers]{cvpr} 

\usepackage{graphicx}
\usepackage{amsmath}
\usepackage{amssymb}
\usepackage{booktabs}

\usepackage{tabularx}
\usepackage{caption}
\usepackage{subcaption}
\usepackage{cite}
\usepackage{url}
\usepackage[hidelinks]{hyperref}

\usepackage[capitalize]{cleveref}
\crefname{section}{Sec.}{Secs.}
\Crefname{section}{Section}{Sections}
\Crefname{table}{Table}{Tables}
\crefname{table}{Tab.}{Tabs.}

\def\cvprPaperID{*****} 
\def\confName{CVPR}
\def\confYear{2022}

\begin{document}

\title{MultiEarth 2022 -- Multimodal Learning for Earth and Environment \\ Workshop and Challenge}

\author{Miriam Cha\textsuperscript{1}, Kuan Wei Huang\textsuperscript{2}, Morgan Schmidt\textsuperscript{2}, Gregory Angelides\textsuperscript{1}, Mark Hamilton\textsuperscript{2}, \\
Sam Goldberg\textsuperscript{2}, Armando Cabrera\textsuperscript{3}, Phillip Isola\textsuperscript{2}, Taylor Perron\textsuperscript{2}, Bill Freeman\textsuperscript{2}, Yen-Chen Lin\textsuperscript{2}, \\
Brandon Swenson\textsuperscript{3}, Jean Piou\textsuperscript{1} \\ \\
\textsuperscript{1}MIT Lincoln Laboratory, [miriam.cha, gregangelides, jepiou]@ll.mit.edu\\
\textsuperscript{2}MIT, [kwhuang, morgansc, markth, sgoldberg, phillipi, perron, billf, yenchenl]@mit.edu \\
\textsuperscript{3}DAF MIT AI Accelerator, [armando.cabrera, brandon.swenson.2]@us.af.mil}


\maketitle

\begin{abstract}
The Multimodal Learning for Earth and Environment Challenge (MultiEarth 2022) will be the first competition aimed at the monitoring and analysis of deforestation in the Amazon rainforest at any time and in any weather conditions. The goal of the Challenge is to provide a common benchmark for multimodal information processing and to bring together the earth and environmental science communities as well as multimodal representation learning communities to compare the relative merits of the various multimodal learning methods to deforestation estimation under well-defined and strictly comparable conditions. MultiEarth 2022 will have three sub-challenges: 1) matrix completion, 2) deforestation estimation, and 3) image-to-image translation. This paper presents the challenge guidelines, datasets, and evaluation metrics for the three sub-challenges. Our challenge website is available at  \url{https://sites.google.com/view/rainforest-challenge}.
\end{abstract}

\section{Introduction}
Despite international efforts to reduce deforestation, the world loses, so far, an area of forest that is equivalent to the size of 40 football fields every minute \cite{wri}.  Deforestation in the Amazon rainforest accounts for the largest share, contributing to reduced biodiversity, habitat loss, and climate change. Since much of the region is difficult to access, satellite remote sensing offers a powerful tool to track changes in the Amazon. However, obtaining a continuous time series of images is hindered by seasonal weather, clouds, smoke, and other inherent limitations of optical sensors. Synthetic aperture radar (SAR), which is insensitive to lighting and weather conditions, appears to be a well suited tool for the task, but SAR images are more difficult for humans to interpret than optical images. A key component of this challenge is to monitor the Amazon rainforest in all weather and lighting conditions using our multimodal remote sensing dataset, which includes a time series of multispectral and SAR images. The 2022 Multimodal Learning for Earth and Environment Challenge (MultiEarth 2022) will be the first competition  aimed at monitoring the Amazon rainforest and predicting deforestation using multimodal representation learning methods.

While considerable research has been devoted to tracking changes in forests \cite{ijgi9100580,lima_2019,Coelho_2021,Ngadze_2020}, the analyses typically rely on passive, optical sensors such as the U.S. Geological Survey’s Landsat, NASA’s Moderate Resolution Imaging Spectroradiometer (MODIS), and the European Space Agency’s Sentinel-2 \cite{sentinel}. These passive, optical sensors need an unobstructed and illuminated view of the scene in order to capture meaningful images. This limits their effectiveness in monitoring the Amazon rainforest with year-round cloud coverage. Beyond the Amazon, approximately 67 percent of Earth’s surface is typically covered by clouds \cite{cloud}. Therefore, using SAR for cloud-free observation will have a broad applicability. 



MultiEarth 2022 will conduct the following sub-challenges to support the interpretation and analysis of the rainforest at any time and any weather conditions:
\begin{itemize}
	\item \textbf{Matrix Completion Sub-Challenge}: given remote sensing  images taken at different locations and dates, and in different modalities, participants are required to predict appearance at a novel [lon, lat, date, modality] query. Performance is measured on the following visual metrics: Peak Signal-to-Noise Ratio (PSNR), Structural Similarity Index Measure (SSIM) \cite{ssim}, Learned Perceptual Image Patch Similarity (LPIPS) \cite{lpips}, and Fr\'{e}chet Inception Distance (FID) \cite{fid}. Detailed challenge description is provided in Section~\ref{subsec:mcc}. 
	\item \textbf{Deforestation Estimation Sub-Challenge}: beyond visual appearance, participants are required to classify whether a region is deforested or not at a novel 
 [lon, lat, date, modality] query. Modality will be `deforestation' for this sub-challenge. Participants will be given the multimodal remote sensing dataset along with the deforestation label maps. Performance is measured on the following metrics: pixel accuracy, F1 score, and Intersection over Union (IoU). Detailed description of this sub-challenge can be found in Section~\ref{subsec:dec}.
	\item \textbf{Image-to-Image Translation Sub-Challenge}: participants are required to predict a set of possible cloud-free corresponding optical images given an input SAR image.  For this sub-challenge, we provide an aligned dataset (e.g. $[\mathbf{x} ,[\mathbf{y}_1,\mathbf{y}_2,...,\mathbf{y}_N]]$) where an input SAR image $\mathbf{x}$ is paired to a set of ground truth optical images $[\mathbf{y}_1,\mathbf{y}_2,...,\mathbf{y}_N]$. Performance is evaluated based on $\sum_j \min_i \Arrowvert f(\mathbf{x})_i - \mathbf{y}_j \Arrowvert$ where $f(\cdot)$ is a prediction of $\mathbf{y}$ given $\mathbf{x}$, and $f$ may make multiple predictions indexed by $i$. Detailed guideline of the sub-challenge is provided in Section~\ref{subsec:i2ic}. 
\end{itemize}

To be eligible to participate in the challenge, every entry has to be accompanied by a paper presenting the results and the methods that created them, which will undergo peer-review. The organizers reserve the right to re-evaluate the findings, but will not participate in the Challenge. 

\begin{figure}[t]
\centering 
\includegraphics[width=.35\textwidth]{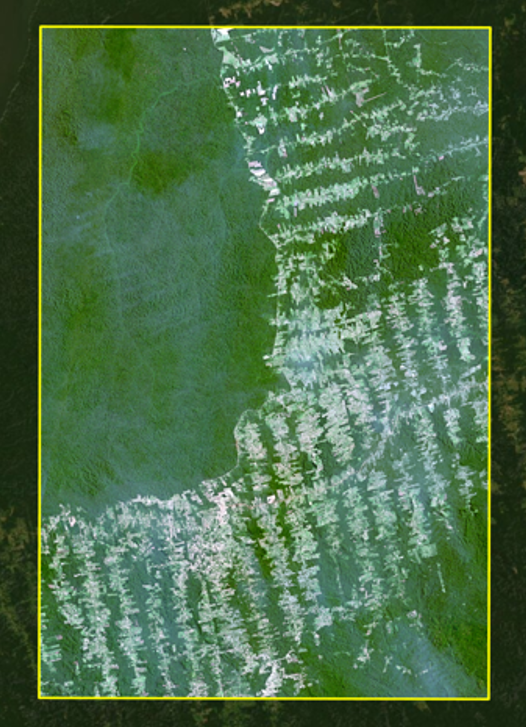} 
\caption{Study area in the Amazon, bounded by (4.39$^{\circ}$ S, 55.2$^{\circ}$ W), (4.39$^{\circ}$ S, 54.48$^{\circ}$ W), (3.33$^{\circ}$ S, 54.48$^{\circ}$ W) and (3.33$^{\circ}$ S, 55.2$^{\circ}$ W).}
\label{fig:tapajos} 
\end{figure}  
\begin{table*}[t]
\small
\centering 
\begin{tabular}{c|c|c|c|c|c}
\hline
\textbf{Sensor} & \textbf{Time} & \textbf{Bands}                                                                                      & \textbf{Resolution (m) } & \textbf{\# Images} & \textbf{Link}\\ \hline \hline
Sentinel-1                     & 2014-2021           & VV, VH                                                                                              & 10                      & 859,627     & \begin{tabular}[c]{@{}c@{}} \href{https://rainforestchallenge.blob.core.windows.net/dataset/sent1_vh_train.zip} {\color{blue}{link1}} \\ \href{https://rainforestchallenge.blob.core.windows.net/dataset/sent1_vv_train.zip}{\color{blue}{link2}}   \end{tabular}   \\ \hline
Sentinel-2                     & 2018-2021           & \begin{tabular}[c]{@{}c@{}}B1, B2, B3, B4, B5,\\ B6, B7, B8, B8A, B9,\\ B11, B12, QA60\end{tabular} & 10                     & 5,395,559    & \begin{tabular}[c]{@{}c@{}} \href{https://rainforestchallenge.blob.core.windows.net/dataset/sent2_qa_b1-b4_train.zip}{\color{blue}{link1}} \\ \href{https://rainforestchallenge.blob.core.windows.net/dataset/sent2_b5-b12_train.zip}{\color{blue}{link2}}   \end{tabular}   \\ \hline
Landsat 5                       & 1984-2012           & \begin{tabular}[c]{@{}c@{}}SR\_B1, SR\_B2,  SR\_B3, \\ SR\_B4, SR\_B5, ST\_B6, \\ SR\_B7, QA\_PIXEL\end{tabular}                    & 30                     & 3,550,368     & \begin{tabular}[c]{@{}c@{}} \href{https://rainforestchallenge.blob.core.windows.net/dataset/landsat5_qa_b1-b3_train.zip}{\color{blue}{link1}} \\ \href{https://rainforestchallenge.blob.core.windows.net/dataset/landsat5_b4-b7_train.zip}{\color{blue}{link2}} \end{tabular}  \\ \hline
Landsat 8                       & 2013-2021           & \begin{tabular}[c]{@{}c@{}}SR\_B1, SR\_B2, SR\_B3, \\ SR\_B4, SR\_B5, SR\_B6, \\ SR\_B7, ST\_B10, QA\_PIXEL\end{tabular}           & 30                     & 2,172,564     & \begin{tabular}[c]{@{}c@{}} \href{https://rainforestchallenge.blob.core.windows.net/dataset/landsat8_qa_b1-b5_train.zip}{\color{blue}{link1}} \\ \href{https://rainforestchallenge.blob.core.windows.net/dataset/landsat8_b6-b10_train.zip}{\color{blue}{link2}}  \end{tabular} \\ \hline
\end{tabular}
\caption{Overview of our multimodal remote sensing dataset that includes Sentinel-1, Sentinel-2, Landsat 5, and Landsat 8.}
\label{tab:data}
\end{table*}

\section{Datasets}
\subsection{Multimodal Remote Sensing Dataset}
\label{subsec:multimodal_data}
Participants will receive a multimodal remote sensing dataset that consists of Sentinel-1, Sentinel-2, Landsat 5, and Landsat 8 as reported in Table~\ref{tab:data}. Sentinel-1 uses a synthetic aperture radar (SAR) instrument, which collects in two polarization bands: VV (vertical transmit/vertical receive) and VH (vertical transmit/horizontal receive). Sentinel-2, Landsat 5, and Landsat 8 use optical instruments, which measure in spectral bands in the visible and infrared spectra. We also include in the dataset the associated layers with cloud quality for Sentinel-2, Landsat 5, and Landsat 8 (i.e. QA60 and QA\_PIXEL). Detailed band designations for each sensor can be found in Google Earth Engine Data Catalog\footnote{\scriptsize Sentinel-1:  \url{https://developers.google.com/earth-engine/datasets/catalog/COPERNICUS_S1_GRD} \\
Sentinel-2: \url{https://developers.google.com/earth-engine/datasets/catalog/COPERNICUS_S2_SR}\\
Landsat 5: \url{https://developers.google.com/earth-engine/datasets/catalog/LANDSAT_LT05_C02_T1_L2} \\
Landsat 8: \url{https://developers.google.com/earth-engine/datasets/catalog/LANDSAT_LC08_C02_T1_L2}
}.  

\begin{figure*}[t]
     \centering
     \begin{subfigure}[b]{0.2\textwidth}
         \centering
         \includegraphics[width=\textwidth]{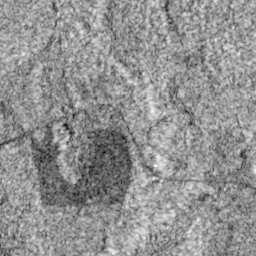}
         \caption{Sentinel-1}
         \label{fig:sentinel1}
     \end{subfigure}
     \hfill
     \begin{subfigure}[b]{0.2\textwidth}
         \centering
         \includegraphics[width=\textwidth]{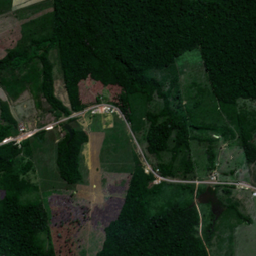}
         \caption{Sentinel-2}
         \label{fig:sentinel2}
     \end{subfigure}
     \hfill
     \begin{subfigure}[b]{0.2\textwidth}
         \centering
         \includegraphics[width=\textwidth]{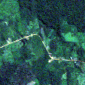}
         \caption{Landsat 5}
         \label{fig:landsat5}
     \end{subfigure}
     \hfill
     \begin{subfigure}[b]{0.2\textwidth}
         \centering
         \includegraphics[width=\textwidth]{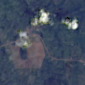}
         \caption{Landsat 8}
         \label{fig:landsat8}
     \end{subfigure}
        \caption{Image examples from our multimodal dataset (LAT/LON: -4.11/-55.14) (a) Sentinel-1 VV on 12/30/2021, (b) Sentinel-2 RGB (B4, B3, B2) on 8/4/2021, (c) Landsat 5 RGB (SR\_B3, SR\_B2, SR\_B1) on 6/12/1998, and (d) Landsat 8 RGB (SR\_B4, SR\_B3, SR\_B2) on 12/4/2021.}
        \label{fig:example_chips}
\end{figure*}

The region of interest is the rectangle bounded by the points (4.39$^{\circ}$ S, 55.2$^{\circ}$ W), (4.39$^{\circ}$ S, 54.48$^{\circ}$ W), (3.33$^{\circ}$ S, 54.48$^{\circ}$ W) and (3.33$^{\circ}$ S, 55.2$^{\circ}$ W) as shown in Figure~\ref{fig:tapajos}. This area is further divided into squares centered on 54 $\times$ 37 = 1998 coordinate pairs by iterating in 0.02$^{\circ}$ increments in the latitude and longitude directions. Our dataset consists of 11,978,118 total training images. (Note that the image-to-image translation sub-challenge will use a subset of the total training images.) Test images are selected from the middle 7 columns (from longitudes -54.78 to -54.90), which are withheld as test region. The training images come from all bands and dates from the 4 sensor collections specified in Table~\ref{tab:data} covering the area outside of the test region. Each image file name is in the format \texttt{\{Collection\}\_\{Band\}\_\{Longitude\}\_} \texttt{\{Latitude\}\_\{Year\}\_\{Month\}\_\{Day\}.tiff}. Sample images of Sentinel-1, Sentinel-2, Landsat 5, and Landsat 8 from our dataset are shown in Figure~\ref{fig:example_chips}.



Satellite images from this dataset are downloaded using Google’s Earth Engine platform. Each coordinate pair is projected from UTM (Universal Transverse Mercator) to EPSG (European Petroleum Survey Group). Then a 256$\times$256 image centered at the coordinate is extracted for Sentinel-1 and Sentinel-2, and a 85$\times$85 image centered at the coordinate is extracted for Landsat 5 and Landsat 8. The difference in pixels accounts for the 10-meter resolution Sentinel-1 and Sentinel-2 vs the 30-meter resolution Landsat 5 and Landsat 8. If the Landsat images are upsampled to 256$\times$256, they will be geospatially aligned with the Sentinel images.

The dataset can be downloaded from the links provided in Table~\ref{tab:data}. Data is freely available for development, research, or educational purposes. For more details, please refer to Google Earth Engine License Agreement\footnote{\url{https://earthengine.google.com/terms/}}. 

\subsection{Deforestation Labels}
\label{subsec:deforestation_data}
\begin{figure*}[t]
     \centering
     \begin{subfigure}[b]{0.21\textwidth}
         \centering
         \includegraphics[width=\textwidth]{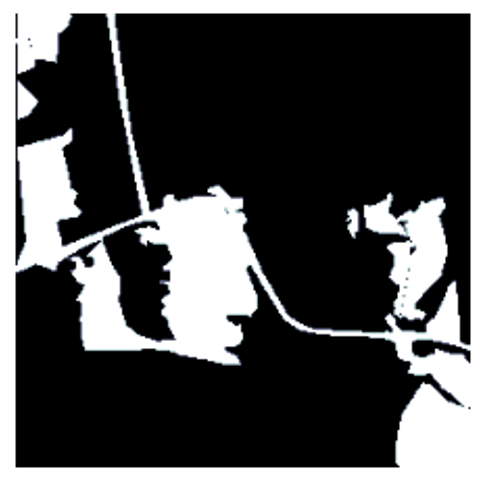}
         \caption{08/2016}
         \label{fig:defor_8_16}
     \end{subfigure}
     \hfill
     \begin{subfigure}[b]{0.21\textwidth}
         \centering
         \includegraphics[width=\textwidth]{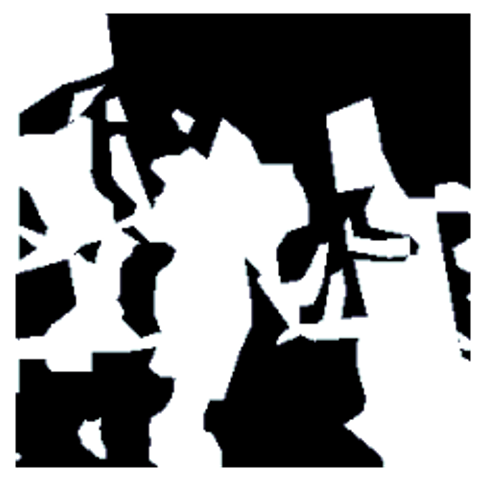}
         \caption{07/2017}
         \label{fig:defor_7_17}
     \end{subfigure}
     \hfill
     \begin{subfigure}[b]{0.21\textwidth}
         \centering
         \includegraphics[width=\textwidth]{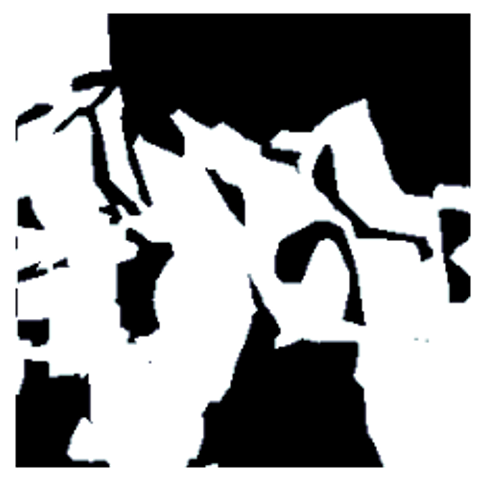}
         \caption{06/2018}
         \label{fig:defor_6_18}
     \end{subfigure}
     \hfill
     \begin{subfigure}[b]{0.21\textwidth}
         \centering
         \includegraphics[width=\textwidth]{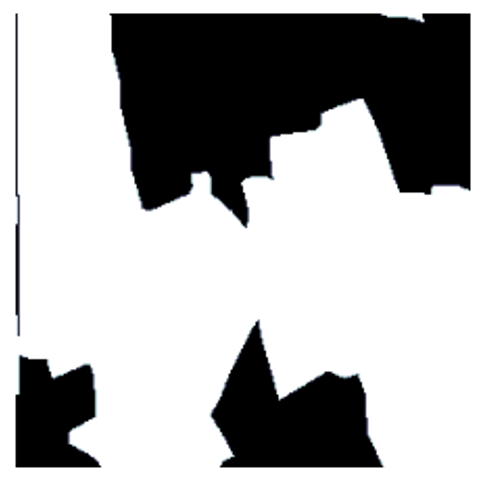}
         \caption{08/2020}
         \label{fig:defor_8_20}
     \end{subfigure}
        \caption{Example images of deforestation label maps corresponding to the location shown in Figure~\ref{fig:example_chips} (LAT/LON: -4.11/-55.14) at various time slices: 08/2016, 07/2017, 06/2018, and 08/2020.}
        \label{fig:example_chips_defor}
\end{figure*}

Labeled data of deforestation for training and testing are manually labeled using cloud-free monthly mosaic satellite images from Planet \cite{planet}. The Planet imagery has 3.7 m spatial resolution and consists of 3 bands (RGB) with images from a single month joined together to achieve an image with the least amount of cloud cover. Most are cloud-free but some have a few clouds. Eleven time slices (months) are labeled over six years, from 2016 to 2021. They include: 08/2016; 07, 08/2017; 06, 08/2018; 07, 08/2019; 06, 08/2020; and 05, 08/2021. The August 2016 image contains a greater number of clouds and artifacts for the mosaic process. August 2021 is labeled by the authors and the remaining ten are labeled by the team at Scale AI \cite{scaleai}. The following guidelines are used by Scale AI to label deforested areas within polygons: 1) areas deforested for human use in any stage of regrowth; 2) forested areas 1 hectare or larger with unbroken forest canopy are not labeled; 3) roads are included as deforested if clearly visible; 4) rivers going through deforested areas are left unlabeled if their area is greater than 1 hectare; 5) clouds or image artifacts are not labeled; 6) snapping was used during labeling to connect polygons. For the August 2021 labeled data, our team does not use snapping but instead leave small spaces between polygons. Therefore, in the labeled dataset, unlabeled areas are mostly forest, but may also contain some clouds or rivers. 
Shapefiles with labeled polygons are converted into rasters with 0’s for forested/other and 1’s for deforested areas. Rasters are divided into georeferenced image chips (256$\times$256 pixels) that correspond with the multimodal remote sensing dataset described in Section~\ref{subsec:multimodal_data}. We show sample images of deforestation label maps at various time slices in Figure~\ref{fig:example_chips_defor}. 

\section{Challenge Tasks}
\subsection{Matrix Completion Sub-Challenge}
\label{subsec:mcc}

\noindent\textbf{Problem Definition.} Most studies of forest monitoring rely on remote sensing data collected by optical  sensors \cite{ijgi9100580,lima_2019,Coelho_2021,Ngadze_2020}. However, it is often difficult, particularly in moist forest areas like the Amazon rainforest, to obtain cloud-free images. This sub-challenge focuses on filling in spatial, temporal, and modality gaps in remote sensing data, especially gaps created by unfavorable lighting, weather conditions, or other atmospheric factors. The Matrix Completion Sub-Challenge represents the sensory world as a huge tensor of measurements, in multiple modalities, sampled at different places and times. This tensor is typically very sparse, with most measurements missing, since satellites only photograph a tiny fraction of all possible wavelengths at all possible positions on the globe at all possible points in time. Participants are required to solve the tensor completion problem on this huge yet sparse multidimensional measurement tensor, and may develop and use methods from representation learning and generative modeling, or may consider other approaches.

\noindent\textbf{Data.} Training set from the multimodal remote sensing dataset described in Section~\ref{subsec:multimodal_data} is provided. 

\noindent\textbf{Metrics.} To rank all submissions to this sub-challenge, imaging results entered by the participants will be evaluated using the following four metrics: Peak Signal-to-Noise Ratio (PNSR), Structural Similarity Index Measure (SSIM) \cite{ssim}, Learned Perceptual Image Patch Similarity (LPIPS) \cite{lpips}, and Fr\'{e}chet Inception Distance (FID) \cite{fid}. 


\noindent\textbf{Submission Format.} For the test input, 2000 test queries will be provided as a list of lists \ie [[${\text{lon}_0, \text{lat}_0, \text{date}_0, \text{modality}_0}],\text{…}, [\text{lon}_{1999}, \text{lat}_{1999}, \text{date}_{1999}$, $\text{modality}_{1999}]]$. Each test query is in the format [lon, lat, date, modality]. For example, $[-55.15, -4.11,  2021\_12\_04,$ $\text{Landsat8\_SR\_B2}]$ will represent Landsat8\_SR\_B2\_-55.15\_-4.11\_2021\_12\_04.tiff. For the test output, participants will submit in total 2,000 256$\times$256 images, one 256$\times$256 image for each input test query. Imagery related to the input test queries will be made available in our website and can be used to help generate the requested output.

\subsection{Deforestation Estimation Sub-Challenge}
\label{subsec:dec}
\noindent\textbf{Problem Definition.} Beyond predicting visual appearance, this sub-challenge is aimed at estimating deforestation from the multimodal remote sensing dataset. As described in the Matrix Completion Sub-Challenge, the multimodal dataset can be represented as a multidimensional data matrix with missing entries. The goal of this sub-challenge is to perform a binary classification to predict whether a region is deforested or not. As solutions for the Matrix Completion Sub-Challenge can be naturally extended,  we strongly encourage participants in the Matrix Completion Sub-Challenge to submit to the Deforestation Estimation Sub-Challenge. 

\noindent\textbf{Data.} Participants will use the multimodal remote sensing dataset described in Section~\ref{subsec:multimodal_data} to predict binary deforestation label maps in Section~\ref{subsec:deforestation_data}.

\noindent\textbf{Metrics.} Performance is measured on the following standard metrics: pixel accuracy, F1 score, and Intersection over Union (IoU). 

\noindent\textbf{Submission Format.} For the test input, 1000 test queries will be provided as a list of lists \ie [[${\text{lon}_0, \text{lat}_0, \text{date}_0, \text{modality}_0}],\text{…}, [\text{lon}_{999}, \text{lat}_{999}, \text{date}_{999}$, $\text{modality}_{999}]]$. Each test query is in the format [lon, lat, date, modality]. For example, $[-55.15, -4.11, 2021\_08\_01,$ $\text{deforestation}]$ will represent deforestation\_-55.15\_-4.11\_2021\_08.png. To have a consistent naming convention for the date (\ie year\_month\_day), we add a nominal day label of “\_01” to all deforestation estimation test queries. For the test output, participants will submit in total 1,000 256$\times$256 binary masks, one 256$\times$256 binary mask for each input test query. Imagery related to the input test queries will be made available in our website and can be used to help generate the requested output.

\subsection{Image-to-Image Translation Sub-Challenge}
\label{subsec:i2ic}

\noindent\textbf{Problem Definition.} Obtaining a continuous time series of view of the Amazon rainforest is hindered by weather, clouds, smoke, and other inherent limitations of passive sensors (e.g. optical sensors) that rely on sunlight. Such limitations produce a major information gap in the Amazon rainforest that gets rain throughout the entire year. Synthetic aperture radar (SAR) is an active sensor that can collect images with relative invariance to weather and lighting conditions. However, visual interpretation of SAR images is not intuitive due to the large dynamic range, low spatial correlation, and radar-specific geometry distortion. To enhance SAR interpretability, this sub-challenge is aimed at modeling a distribution of possible electro-optical (EO) image outputs conditioned on a SAR input image. Here, EO image is a 3-channel RGB image from Sentinel-2. In Sentinel-2, RGB bands are represented as B4, B3, and B2, respectively. SAR image is a 2-channel Sentinel-1 image consisting of VV and VH bands. In this Image-to-Image Translation Sub-Challenge, participants need to model a distribution of potential results in a conditional generative modeling setting.

\noindent\textbf{Data.} For this sub-challenge, we provide JSON files specifying which Sentinel-2 EO images (B4, B3, and B2) correspond to which SAR images. We provide two JSON files: one for Sentinel-1 VV band\footnote{\url{https://rainforestchallenge.blob.core.windows.net/dataset/sentinel_vv_image_alignment_train.json}} and another for Sentinel-1VH band \footnote{\url{https://rainforestchallenge.blob.core.windows.net/dataset/sentinel_vh_image_alignment_train.json}}. The mappings in both files are identical. The aligned dataset will have the following format: $[\mathbf{x} ,[\mathbf{y}_1,\mathbf{y}_2,...,\mathbf{y}_N]]$ where a SAR image $\mathbf{x}$ is paired to a set of ground truth EO images $[\mathbf{y}_1,\mathbf{y}_2,...,\mathbf{y}_N]$. For each SAR image, all EO images of the same geographic region and which were collected within 7 days of the SAR image timestamp will be identified.  On average 3 EO images are paired with each SAR image ($N \approx 3$). 

\noindent\textbf{Metrics.} Performance is evaluated based on following:
\begin{equation}
\sum_j \min_i \Arrowvert f(\mathbf{x})_i - \mathbf{y}_j \Arrowvert
\end{equation}
 where $f(\mathbf{x})_i$ is a set of possible EO images translated from  a generative model $f(\cdot)$ conditioned on an input SAR image $\mathbf{x}$, and  $\mathbf{y}_j$ is an EO image from the corresponding ground truth set. This metric also evaluates diversity of generated output images staying faithful to the diversity of the ground-truth EO data. 

\noindent\textbf{Submission Format.} For testing, we will provide 5000 Sentinel-1 SAR images, where each SAR image is 256$\times$256$\times$2 and the 2 channels correspond to the Sentinel-1 VV and VH bands, respectively. For the test output, participants will submit in total 15,000 256$\times$256$\times$3 EO images, three translated 256$\times$256$\times$3 EO images for each input Sentinel-1 SAR image. The EO image channels correspond to the Sentinel-2 RGB channels (B4, B3, B2). This is a multimodal image-to-image translation problem where participants will generate three possible EO images given an input SAR image.

\section{Conclusion}
We introduce MultiEarth 2022 – the first open Multimodal Learning for Earth and Environment Challenge. It comprises three sub-challenges: 1) matrix completion, 2) deforestation estimation, and 3) image-to-image translation. This manuscript describes MultiEarth 2022’s challenge conditions, data, evaluation metrics, and submission guideline. Only a few labeled, multimodal datasets including passive and active sensors have been publicly available, limiting the number of participants who can research and analyze Earth’s surface at all times and in all weather conditions. We collect and disseminate a multimodal dataset that includes a continuous time series of Sentinel-1, Sentinel-2, Landsat 5 and Landsat 8, with deforestation labels. 

\footnotesize
\section{Acknowledgments}
The authors would like to thank Dr. M.K. Newey for helping the team labeling the August 2021 time slice data, and Scale AI, for its efforts to label, in a short period of time, the remaining ten time slices considered in this MultiEarth2022.

Research was sponsored by the United States Air Force Research Laboratory and the United States Air Force Artificial Intelligence Accelerator and was accomplished under Cooperative Agreement Number FA8750-19-2-1000. The views and conclusions contained in this document are those of the authors and should not be interpreted as representing the official policies, either expressed or implied, of the United States Air Force or the U.S. Government. The U.S. Government is authorized to reproduce and distribute reprints for Government purposes notwithstanding any copyright notation herein.

{\footnotesize
\bibliographystyle{ieee_fullname}
\bibliography{paper}
}

\end{document}